# On-ramp and Off-ramp Traffic Flows Estimation Based on A Data-driven Transfer Learning Framework

**Xiaobo Ma, Ph.D.**
Department of Civil & Architectural Engineering & Mechanics
The University of Arizona
1209 E 2nd St, Tucson, AZ 85721
Email: xiaoboma@arizona.edu

**Abolfazl Karimpour, Ph.D.**
College of Engineering
State University of New York Polytechnic Institute
100 Seymour Rd, Utica, NY 13502
Email: karimpa@sunypoly.edu

**Yao-Jan Wu, Ph.D., P.E.**
Department of Civil & Architectural Engineering & Mechanics
The University of Arizona
1209 E 2nd St, Tucson, AZ 85721
Email: yaojan@arizona.edu

**ABSTRACT**
To develop the most appropriate control strategy and monitor, maintain, and evaluate the traffic performance of the freeway weaving areas, state and local Departments of Transportation need to have access to traffic flows at each pair of on-ramp and off-ramp. However, ramp flows are not always readily available to transportation agencies and little effort has been made to estimate these missing flows in locations where no physical sensors are installed. To bridge this research gap, a data-driven framework is proposed that can accurately estimate the missing ramp flows by solely using data collected from loop detectors on freeway mainlines. The proposed framework employs a transfer learning model. The transfer learning model relaxes the assumption that the underlying data distributions of the source and target domains must be the same. Therefore, the proposed framework can guarantee high-accuracy estimation of on-ramp and off-ramp flows on freeways with different traffic patterns, distributions, and characteristics. Based on the experimental results, the flow estimation mean absolute errors range between 23.90 veh/h to 40.85 veh/h for on-ramps, and 31.58 veh/h to 45.31 veh/h for off-ramps; the flow estimation root mean square errors range between 34.55 veh/h to 57.77 veh/h for on-ramps, and 41.75 veh/h to 58.80 veh/h for off-ramps. Further, the comparison analysis shows that the proposed framework outperforms other conventional machine learning models. The estimated ramp flows based on the proposed method can help transportation agencies to enhance the operations of their ramp control strategies for locations where physical sensors are not installed.

# 1. Introduction

Freeways are the main arteries of every transportation network and usually create the highest level of mobility between every pair of origin and destination (Guan & He, 2008). Freeways are also connected to the lower mobility and higher accessibility arterials via on-ramps and off-ramps. A major traffic flow on freeways uses the off-ramps to get access to arterials, and a major traffic flow on arterials uses the on-ramps to get into the freeways (Nohekhan et al., 2021). Road users expect to move along their daily trips with the highest level of speed and reliability(Meng et al., 2016; Rubaiyat et al., 2018; Z. Zhong et al., 2020). However, due to recurrent and nonrecurrent traffic congestion occurring across the freeway and at entering/exiting locations, this is not always feasible (Papageorgiou & Kotsialos, 2002). The high volume of vehicles entering and exiting the freeway during peak periods could significantly deteriorate downstream traffic conditions as well as mainline freeways. Therefore, in order to actively alleviate the traffic congestion on the mainline and ramps, fully utilize the freeway capacity, and monitor the performance of the roads, freeways should be equipped with appropriate control strategies that can provide smooth traffic progression across the freeway (Ma et al., 2020a).

The accurate estimation of traffic flows at on-ramps and off-ramps holds critical significance for Federal, State, and Local agencies(Luo et al., 2022; Ma, 2022; Ma et al., 2020b). It serves as the foundation for developing highly effective traffic control strategies and enables meticulous monitoring, maintenance, and evaluation of the traffic performance within freeway weaving areas (Daganzo, 2002; Liu et al., 2017). Furthermore, on-ramp and off-ramp traffic flows play pivotal roles in sophisticated traffic modeling and accurate traffic state estimation (Wang et al., 2022; Zhang et al., 2022). However, traffic flows of on-ramps and off-ramps are not always readily available to transportation agencies. That is, not all ramps are equipped with data collection sensors (such as loop detectors) for data collection, and/or they might not have designated traffic messaging channels for data communication (Nohekhan et al., 2021). Usually, transportation agencies are facing two major types of missing traffic data at on-ramps and off-ramps: 1) instances where continuous communication loss happens, which could lead to complete or partial missing data, and 2) instances where no physical sensor is installed for data collection. Complete or partial missing traffic data can severely impact the data quality and can reduce the effectiveness of traffic system management programs (Duan et al., 2016; Karimpour et al., 2019; Tang et al., 2015). The unavailability of data due to a lack of sensors can as well negatively impact the whole network modeling since traffic flow variables are a requirement for freeway network design and modeling (Kan et al., 2021).

In order to deal with the missing traffic data collected from the loops at on-ramps and off-ramps, missing values could be imputed (Karimpour et al., 2019). Data imputation techniques have been widely applied for imputing missing values for different traffic variables, including missing ramp data (Z. Liu et al., 2008). Common techniques to impute the missing traffic data are categorized as prediction, interpolation, and statistical learning techniques (Duan et al., 2016; Y. Li et al., 2014). Using historical data, autoregressive moving average (ARIMA), and pattern-matching algorithms are some of the prediction techniques that have been widely used for traffic data imputation (Gan et al., 2015; L. Li et al., 2015; M. Zhong et al., 2006). Interpolation techniques are very similar to prediction techniques. For Interpolation, the missing values are substituted with values based on neighboring data or historical data (Z. Liu et al., 2008). For instance, Liu et al. used K-Nearest neighbors for imputing the missing flow data during holiday periods (Z. Liu et al., 2008). Machine learning (Ran et al., 2015), deep learning (Xie et al., 2010), principal component analysis (PCA) (L. Li et al., 2013; L. Qu et al., 2009), and random forest (Ou et al., 2017) models are some types of statistical learning techniques (Duan et al., 2016). Comparably, statistical learning techniques are more accurate, flexible, and can accurately distinguish the interaction among different data points (Duan et al., 2016; Y. Li et al., 2014).

Comparably, less effort has been done to estimate missing traffic data at on-ramps and off-ramps in locations where no physical sensors have been installed (Kan et al., 2021; Nohekhan et al., 2021). Traditional approaches mainly focused on the estimation of link flow using the origin-destination (OD) matrix. However, gathering up-to-date OD is not always plausible (Marzano et al., 2009). More recent studies have focused on macroscopic traffic flow modeling for estimating link flow. However, macroscopic traffic flow modeling is very time-consuming, not always accurate, and requires extensive model calibration (Nohekhan et al., 2021). Machine learning algorithms have been applied extensively to solve different problems(Y. Yang et al., 2023; Zhao et al., 2022). Recent advances in the area of Artificial Intelligence (AI) and Machine Learning (ML) have allowed researchers and practitioners to use machine learning techniques, such as autoregressive integrated moving average(S. He et al., 2023), random forest(X. Qu & Hickey, 2022), support vector machine(R. Wang & Qu, 2022), artificial neural network(Y. He, Tian, et al., 2023; Y. He, Zhang, et al., 2023; J. Wu et al., 2023), convolutional neural network(Cui et al., 2022; Hong et al., 2022; Sun et al., 2022; X. Yang et al., 2023; Yi & Qu, 2022; Zhan et al., 2022; D. Zhang et al., 2023), recurrent neural network(Hu et al., 2021), generative adversarial network(Z. Huang et al., 2020; Z. R. Huang & Chiu, 2020; X. Li et al., 2022), reinforcement learning(J. Chen et al., 2022; Z. Chen et al., 2022; Cheng et al., 2021; W. Liu et al., 2023; Mei et al., 2023; P. Wang et al., 2021; Zhou et al., 2022), curriculum learning(Dou et al., 2023), transfer learning(W. Liu et al., 2022), contrastive learning(Pokle et al., 2022; Shen et al., 2021), representation learning(Dou, Jia, et al., 2022; Dou, Pan, et al., 2022), federated learning(X. Wu et al., 2023), incremental learning(You et al., 2022), transformer(D. Zhang & Zhou, 2023), natural language processing(Yan et al., 2023) to finish various tasks. Also, the use of machine learning techniques for solving complex problems in the transportation domain has gained in popularity in recent years(L. Zhang & Lin, 2022). For instance, Nohekhan et al. (2021) used a deep learning approach to estimate the hourly traffic volume of off-ramps (Nohekhan et al., 2021). In another study, Kan et al. (2021) used random forest (RF) and gradient boosting (GBM) to estimate the ramp pairs on a Shanghai Urban Expressway and an Intercity Highway in California (Kan et al., 2021).

There remain two main limitations while trying to estimate missing traffic data at on-ramps and off-ramps where no physical sensors have been installed using emerging AI and ML-based approaches. First, most of the proposed estimation models in the literature require extensive data collection. However, some of these data might not be readably available or needed to be separately estimated from Highway Capacity Manual guidelines. For instance, in the model proposed by Nohekhan et al. (2021), the model required average annual daily traffic, road classification, and segment free-flow speed as their input variables, and estimated the traffic volume of off-ramps in an hourly aggregation level (Nohekhan et al., 2021). Similarly, in the model proposed by Kan et al. (2021), in order to train the estimation model, the authors used the flow capacity of the weaving area between the on-ramp and the off-ramp, on-ramp capacity, and peak on-ramp demand (Kan et al., 2021). Second, most of the estimation models in the literature are generic models, considering different freeway segments have different traffic patterns, models built based on certain locations may not perform well when being applied to other locations (Muralidharan & Horowitz, 2009).

This study aims to tackle the aforementioned limitations by proposing a data-driven transfer learning framework. The proposed framework accurately estimates missing traffic flows at on-ramps and off-ramps by solely using data collected from loop detectors on freeway mainlines. In addition, the proposed framework develops scene-specific models by employing a transfer learning model. Because the transfer learning model relaxes the assumption that the underlying data distributions of the source and target domains must be the same, the proposed framework can guarantee high-accuracy estimation of on-ramp and off-ramp flows on freeways with different traffic patterns, distributions, and characteristics. The performance of the proposed framework is evaluated using freeway traffic data collected from the South-Bound stretch of State Route 51 (SR 51) and Loop 101 (L101) in the Phoenix Metropolitan area.

## 2. Problem Definition and Preliminaries
### 2.1. Notations

Figure 1 lists all the variables and their definitions that were used for the model development.

Table 1 Notations and Variables

| Variables | Definition |
|---|---|
| $q_i^{up}$ | Upstream traffic flow rate at the time interval $i$ |
| $\tilde{q}_i^{up}$ | Upstream traffic flow-related statistical parameter at the time interval $i$ |
| $\tilde{o}_i^{up}$ | Upstream occupancy-related statistical parameter at the time interval $i$ |
| $\tilde{v}_i^{up}$ | Upstream traffic speed-related statistical parameter at the time interval $i$ |
| $q_i^{down}$ | Downstream traffic flow rate at the time interval $i$ |
| $\tilde{q}_i^{down}$ | Downstream traffic flow-related statistical parameter at the time interval $i$ |
| $\tilde{o}_i^{down}$ | Downstream occupancy-related statistical parameter at the time interval $i$ |
| $\tilde{v}_i^{down}$ | Downstream traffic speed-related statistical parameter at the time interval $i$ |
| $q_i^{on}$ | On-ramp traffic flow rate at the time interval $i$ |
| $q_i^{off}$ | Off-ramp traffic flow rate at the time interval $i$ |

Figure 1 provides a freeway stretch that includes an upstream segment, a downstream segment, an on-ramp segment, and an off-ramp segment. The tilde symbol ~ indicates the statistical parameter at the time interval $i$ from a certain segment. For example, $\tilde{v}_i^{up}$ represents the upstream traffic speed-related parameters at the time interval $i$ (i.e., $\tilde{v}_i^{up}$ could refer to any upstream traffic speed parameters, such as mean speed, speed variance, speed standard deviation, speed kurtosis, or speed skewness).

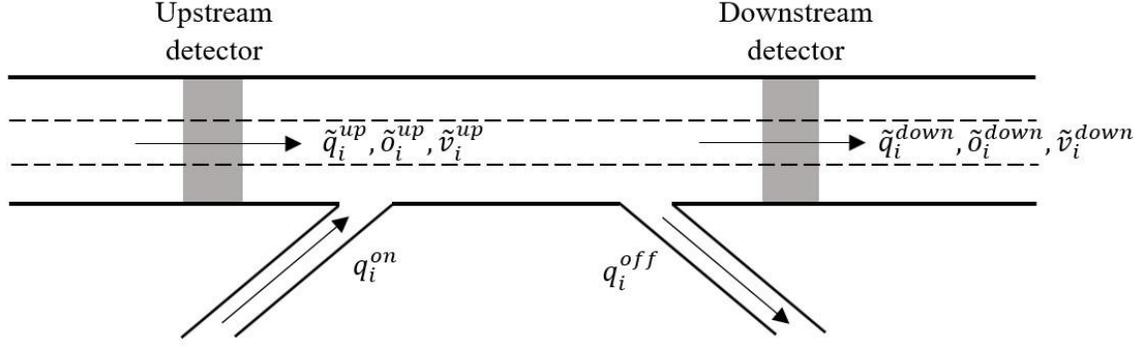
Figure 1 Illustration of a qualitative freeway stretch

## *2.2. Parameter definition and data encoding*
This subsection provides detailed formula used for estimating the traffic variables listed in Table 1. Let $R_n:\{r_1, ..., r_n\}$ denotes a time series within $i$ time interval.
(1) Arithmetic Mean Definition
The arithmetic mean is the sum of the sample values divided by the number of items in the time series. The arithmetic mean is given by
$$\bar{r} = \frac{1}{n} * \sum_{l=1}^{n} r_l \quad (r_1 \leq r_l \leq r_n) \tag{1}$$
(2) Variance Definition
Variance measures how spread out the sample values are around the mean. Variance is given by
$$\sigma^2 = \frac{\sum_{l=1}^{n}(r_l - \bar{r})^2}{n-1} \quad (r_1 \leq r_l \leq r_n) \tag{2}$$
(3) Standard Deviation Definition
Standard deviation measures the amount of variation in the time series. The standard deviation is given by
$$\sigma = \sqrt{\frac{\sum_{l=1}^{n}(r_l - \bar{r})^2}{n-1}} \quad (r_1 \leq r_l \leq r_n) \tag{3}$$
(4) Kurtosis Definition
Kurtosis describes the flatness of the probability distribution of sample values. In this study, kurtosis is calculated by
$$\kappa = \frac{1}{n} * \sum_{l=1}^{n} \left(\frac{r_l - \bar{r}}{\sigma}\right)^4 - 3 \quad (r_1 \leq r_l \leq r_n) \tag{4}$$
Where $\bar{r}$ is the mean of the distribution, and $\sigma$ represents the standard deviation.
(5) Skewness Definition
Skewness is a measure of the asymmetry of the probability distribution of sample values. Skewness is defined as
$$s = \frac{1}{n} * \sum_{l=1}^{n} \left(\frac{r_l - \bar{r}}{\sigma}\right)^3 \quad (r_1 \leq r_l \leq r_n) \tag{5}$$
Where $\bar{r}$ is the mean of the distribution, and $\sigma$ represents the standard deviation.
(6) Minute-of-hour, hour-of-day, and day-of-week
Each hour has four 15-minute intervals; therefore, the minute-of-hour (MOH) takes values from 1 to 4 to represent each 15-minute interval. Hour-of-day (HOD) is represented by subsequent numbers, starting from 0 to 23 (0 representing midnight, and 23 representing 11 pm). Day-of-week (DOW) is represented by a number ranging from 1 to 7, with Sunday encoded as 1 and Saturday encoded as 7.

## 2.3. Problem formulation

The purpose of this study is to build a machine-learning framework to estimate on-ramp and off-ramp traffic flows for locations where physical sensors are not installed. To achieve this goal, this study employs a transfer learning model. Transfer learning is a supervised boosting technique that effectively utilizes the valuable knowledge acquired from trained models. It involves storing this knowledge and then applying it to a separate yet similar model, leading to innovative solutions for related problems. Specifically, this study trains models on a known freeway location (henceforth referred to as "source domain") and transfers the well-trained models to estimate on-ramp and off-ramp flows for a new freeway location (henceforth referred to as "target domain"). As shown in Eq. (6) and Eq. (7), by associating spatial-temporal traffic variables for a known freeway location, the learned function $F_1(\cdot)$ and $F_2(\cdot)$ can be used for traffic flow estimation on a new freeway location.

$$F_1([q_i^{up}, \tilde{q}_i^{up}, \tilde{o}_i^{up}, \tilde{v}_i^{up}, q_i^{down}, \tilde{q}_i^{down}, \tilde{o}_i^{down}, \tilde{v}_i^{down}]) = [q_i^{on}] \quad (6)$$

$$F_2([q_i^{up}, \tilde{q}_i^{up}, \tilde{o}_i^{up}, \tilde{v}_i^{up}, q_i^{down}, \tilde{q}_i^{down}, \tilde{o}_i^{down}, \tilde{v}_i^{down}]) = [q_i^{off}] \quad (7)$$

Given source domain data $D_S = \{(x_{S_1}, y_{S_1}), \ldots, (x_{S_n}, y_{S_n})\}$, where $x_{S_i} \in \mathcal{X}_S$ is the data instance and $y_{S_i} \in \mathcal{Y}_S$ is the corresponding label. $F_1(\cdot)$ and $F_2(\cdot)$ can be generalized as $F_3(\cdot)$.

$$F_3([\mathcal{X}_S]) = [\mathcal{Y}_S] \quad (8)$$

Let target domain data $D_T = \{(x_{T_1}, y_{T_1}), \ldots, (x_{T_k}, y_{T_k})\}$, where $x_{T_i} \in \mathcal{X}_T$ is the data instance and $y_{T_i} \in \mathcal{Y}_T$ is the corresponding label. Assume $\mathcal{X}_T'$ and $\mathcal{Y}_T'$ are small portions of data from the target domain, $\mathcal{X}_S'$ and $\mathcal{Y}_S'$ are small portions of data from the source domain. The transfer learning model employed in this study needs $\mathcal{X}_T'$ and $\mathcal{Y}_T'$ to be added to the model training process to fine-tune $F_3(\cdot)$ to acquire $F_4(\cdot)$.

In this case, $F_4(\cdot)$ could have satisfactory performance for the target domain.

$$F_4([\mathcal{X}_S, \mathcal{X}_T']) = [\mathcal{Y}_S, \mathcal{Y}_T'] \quad (9)$$

Considering $\mathcal{X}_T'$ and $\mathcal{Y}_T'$ are not available in this study, $\mathcal{X}_S'$ and $\mathcal{Y}_S'$ are extracted from the source domain to substitute for $\mathcal{X}_T'$ and $\mathcal{Y}_T'$ to form $F_5(\cdot)$. Detailed explanation on how to select $\mathcal{X}_S'$ and $\mathcal{Y}_S'$ are illustrated in section 3.4. In the end, $F_5(\cdot)$ is built for on-ramp and off-ramp traffic flow estimation in this study.

$$F_5([\mathcal{X}_S, \mathcal{X}_S']) = [\mathcal{Y}_S, \mathcal{Y}_S'] \quad (10)$$

## 3. Methodology
### 3.1. Research framework

The main idea behind the proposed framework is to estimate on-ramp and off-ramp traffic flows for the location where no physical sensor is installed. The proposed framework is a stepwise procedure for estimating traffic flows. In the first step, traffic variables from upstream and downstream of freeway mainlines are extracted. Let's assume the whole data set of the source and target domains corridor(s) are denoted as $\mathbb{D}_S$ and $\mathbb{D}_T$, respectively; in this case $\mathbb{D}_S = (\mathbb{X}_S, \mathbb{Y}_S)$ and $\mathbb{D}_T = (\mathbb{X}_T, \mathbb{Y}_T)$. Both $\mathbb{D}_S$ and $\mathbb{D}_T$ have two parts: data instances $\mathbb{X}_S$ and $\mathbb{X}_T$ as well as labels $\mathbb{Y}_S$ and $\mathbb{Y}_T$. $\mathbb{X}_S$ and $\mathbb{X}_T$ contain all the variables extracted from the first step, assume the number of variables extracted is $p$, then $\mathbb{X}_S$ and $\mathbb{X}_T$ can be regarded as $n$ by $p$ matrices.

In the second step, ridge regression is used for, a) interpreting the relationship between the on-ramp and off-ramp flows and the traffic variables of the upstream and downstream segments, and b) important variable selection. With the important variable selected, an inter-corridor similar-location matching algorithm is proposed to match each location of the target domain to a certain location of the source domain. Let's assume the number of selected variables is $q$. Let $D_T$ represent data from the target domain location and $D_S$ represent data from matched source domain location. Similarly, $D_S = (\mathcal{X}_S, \mathcal{Y}_S)$ and $D_T = (\mathcal{X}_T, \mathcal{Y}_T)$. Then both $\mathcal{X}_S$ and $\mathcal{X}_T$ become $n$ by $q$ matrices. Before proceeding with the last step, target domain data substitution is conducted to extract $\mathcal{X}_S'$ and $\mathcal{Y}_S'$ from the source domain location to substitute for $\mathcal{X}_T'$ and $\mathcal{Y}_T'$. The Ridge regression procedure is presented in section 3.2.

Next, a transfer learning model named Two-stage TrAdaBoost.R2 (henceforth referred to as TrA) is deployed to estimate traffic flows for all the locations' on-ramps and off-ramps in the target domain corridor. Finally, to evaluate the model estimation accuracy, the model developed for the source domain is used to estimate the missing variables in the target domain.

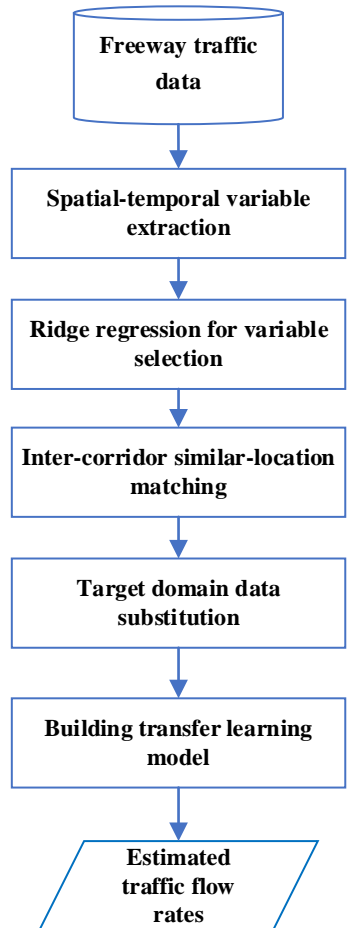

Figure 2 Research framework

## 3.2. Ridge regression

In this study, ridge regression is used for variable importance selection and spatial-temporal traffic variables interpretation. Ridge regression is a supplement to least squares regression that trades unbiasedness in exchange for high numerical stability, thereby obtaining higher calculation accuracy (Hoerl & Kennard, 2000). Assuming $X$ is a $n$ by $p$ matrix with centered columns, $Y$ is a centered $n$ vector. Ridge regression estimator $\beta_{ridge}$ is chosen to minimize the penalized sum of squares.

$$\sum_{i=1}^{n}(y_i - \sum_{j=1}^{p} x_{ij}\beta_j)^2 + \lambda \sum_{j=1}^{p} \beta_j^2 \tag{11}$$

The solution to the minimization problem is

$$\beta_{ridge} = (X^T X + \lambda I)^{-1} X^T Y \tag{12}$$

Where $I$ is the identity matrix, $\lambda$ is selected to give the minimum mean cross-validated error.

## 3.3. Inter-corridor similar-location matching

After choosing the important $q$ traffic variables using ridge regression, the next step is to find a matching function $\mathcal{M}: D_T \to D_S$ to map each location of the target corridor to a certain location of the source corridor. The objective is to find a source location that has a similar spatial-temporal pattern to the target location. To do so, the Pearson Correlation test is calculated pairwisely between the same traffic variables from the source and target locations. Finally, for each target location, the source location with the largest correlation value is selected.

Let's define $\alpha_{S,i}$ and $\alpha_{T,i}$ as the $i$-th traffic variable from source locations and target locations. The matched source location for each target location needs to meet the following criteria:

$$max \sum_{i=1}^{q}(corr(\alpha_{S,i}, \alpha_{T,i})) \tag{13}$$

## 3.4. Target domain data substitution

Recall source domain data $D_S = \{(x_{S_1}, y_{S_1}), \ldots, (x_{S_n}, y_{S_n})\}$, where $x_{S_i} \in \mathcal{X}_S$ is the data instance and $y_{S_i} \in \mathcal{Y}_S$ is the corresponding label. Target domain data is denoted as $D_T = \{(x_{T_1}, y_{T_1}), \ldots, (x_{T_k}, y_{T_k})\}$, where $x_{T_i} \in \mathcal{X}_T$ is the data instance and $y_{T_i} \in \mathcal{Y}_T$ is the corresponding label. When applying TrA to induce a predictive model, some labeled data in the target domain are required. To meet this requirement, $D_S' = \{(x_{S_j}, y_{S_j})\}$ ($D_S' \subsetneqq D_S, 1 \leq j \leq m < n$) is created to act as a substitute for the labeled data in the target domain. $D_S'$ only contains pairs $(x_{S_j}, y_{S_j})$ in which the cosine similarity of $x_{S_j}(1 \leq j \leq m < n)$ and $x_{T_i}(1 \leq i \leq k)$ is less than or equal to a threshold value $\vartheta$. $\vartheta$ is set as a hyper-parameter to select the top 10% of data instances that have the highest similarity to $\mathcal{X}_T$ from $\mathcal{X}_S$ (Dai et al., 2007).

## 3.5. Two-stage TrAdaBoost.R2

As one of the instances-based inductive transfer learning methods, the TrA is a boosting-based algorithm that aims to increase the prediction performance by linearly combining the weak estimators and forming stronger estimators (Pardoe & Stone, 2010; Yehia et al., 2021).

$$F(x) = \sum_{t=1}^{T} \beta_t G(x, \gamma_t) \tag{14}$$

Here $\beta_t$ is the weight of the weak estimator, $G(x, \gamma_t)$ is the weak estimator, and $\gamma_t$ is the optimal parameter of the weak estimator.

Let $D_S$ (of size $n$) denotes labeled source domain data and $D'_S$ (of size $m$) denotes labeled data in the target domain. $D$ is the combination of $D_S$ and $D'_S$. $S$ is the number of steps, $F$ is the number of folds for cross-validation. $S$ and $F$ are set as 10 and 5, respectively. The TrA algorithm is shown below.

**Input** $D$, $S$, and $F$. Setting the initial weight vector $\mathbf{w}^1$ such that $\omega_i^1 = \frac{1}{n+m}$ for $1 \leq i \leq n+m$

**For** $t = 1, \ldots, S$:
1. Call AdaBoost.R2 (Drucker, 1997) with $D$, estimator $G(x)$, and weight vector $\mathbf{w}^t$. $D_S$ is unchanged in this procedure. Obtaining an estimate $error_t$ of $model_t$ using $F$-fold cross-validation.
2. Call estimator $G(x)$ with $D$ and weight vector $\mathbf{w}^t$
3. Calculate the adjusted error $e_i^t$ for each instance as in AdaBoost.R2
4. Update the weight vector

$$\omega_i^{t+1} = \begin{cases} \frac{\omega_i^t \beta_t^{e_i^t}}{Z_t}, & 1 \leq i \leq n \\ \frac{\omega_i^t}{Z_t}, & n+1 \leq i \leq n+m \end{cases}$$

where $Z_t$ is a normalizing constant, and $\beta_t$ is chosen such that the weight of the target (final $m$) is $\frac{m}{n+m} + \frac{t}{(S-1)}(1 - \frac{m}{n+m})$

**Output** $F(x) = model_t$, where $t = \mathrm{argmin}_i error_i$

## 4. Case Study
### *4.1. Data description*
Figure 3 illustrates the layout of the study corridors in the State of Arizona. In total, nine pairs of on-ramp and off-ramp (in red) along State Route 51 (SR51) and seven pairs of on-ramp and off-ramp (in blue) along Loop 101 (L101) were selected for this study. Loop detector data collected from SR51 were chosen as the source domain, and Loop detector data from L101 were used for the target domain. It is worth mentioning that, to imitate locations where no physical sensors were installed, the on-ramp and off-ramp flows on the target domain (L101) were assumed to be missing. The loop detector data consists of 20-second speed, volume, and occupancy data collected from dual-loop detectors installed on the mainlines and on-ramps of freeways. Since the high occupancy (HOV) lanes have lower traffic volumes and higher speeds than the general-purpose lanes (Ma et al., 2020a), they were excluded from this study. The numbers in Figure 3 represent the on-ramp and off-ramp corresponding loop detector numbers; red numbers for the source domain, and blue numbers for the target domain.

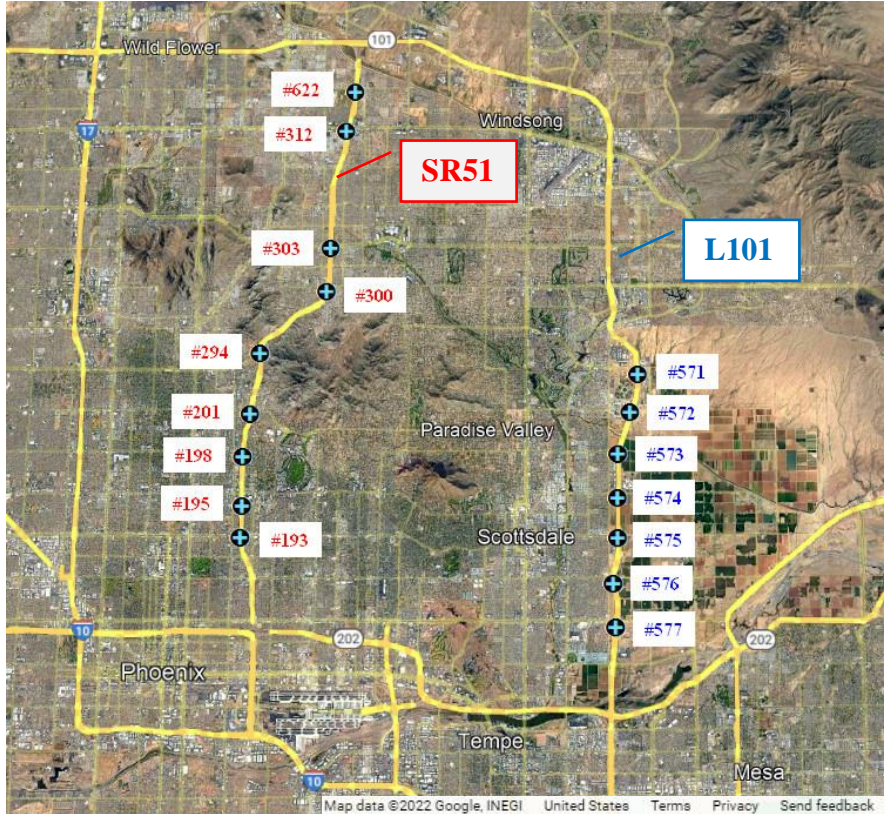
Figure 3 Layout of the study corridor

## *4.2. Experiment design*
### *4.2.1. Data sample*
In order to validate the performance of the proposed framework, one month (2019-07-31 ~ 2019-08-31) of loop detector data was collected from the study sites. The raw data (traffic volume, speed, and occupancy) that was collected from the loop detectors had an aggregation level of 20 seconds. To process the raw data, summed traffic volume, volume-weighted speed, and volume-weighted occupancy were calculated for each segment. Then, the traffic variables listed in section 2.2 were calculated based on 15-minute intervals for the upstream, downstream, on-ramp, and off-ramp segments. Notably, each location (upstream, downstream, on-ramp, and off-ramp segment) have 3,072 (32 days * 24 hours *4) data samples.

### *4.2.2. Model structure and hyper-parameters*
Four representative machine learning models, k-nearest neighbor (KNN), support vector regression (SVR), artificial neural network (ANN), and extreme gradient boosting (XGB), were selected as the baseline models to examine the feasibility of the proposed framework for estimating on-ramp and off-ramp traffic flows. Since the hyperparameter values can greatly influence the estimation accuracy of each machine learning model, initially a grid search approach was developed to find the optimal values for the hyperparameters. The results of the grid search for the optimal structures for different models and tasks are tabulated in Table 2.

Table 2 Grid Search Results: Optimal Values of Hyperparameters

| Model | Test range | Upstream speed prediction | Downstream speed prediction |
|---|---|---|---|
| KNN | n_neighbors [1,50] | n_neighbors=30 | n_neighbors=30 |
| SVR | gamma [0.1,1,10]<br>C [0.1,1,10,100] | gamma=0.1<br>C=10 | gamma=0.1<br>C=10 |
| ANN | hidden_layer_sizes [50,100,150,200,250,300] | hidden_layer_sizes= (100,150) | hidden_layer_sizes= (100,150) |
| XGB | learning_rate [0.0001, 0.001, 0.01, 0.1]<br>max_depth [1,30]<br>n_estimators [50,100,150,200,250,300] | learning_rate=0.01<br>max_depth=5<br>n_estimators=200 | learning_rate=0.01<br>max_depth=5<br>n_estimators=200 |
| TrA | learning_rate [0.0001, 0.001, 0.01, 0.1]<br>max_depth [1,30]<br>n_estimators [50,100,150,200,250,300] | learning_rate=0.1<br>max_depth=20<br>n_estimators=200 | learning_rate=0.1<br>max_depth=20<br>n_estimators=200 |

*4.2.3. Measurements of Effectiveness*

Mean Absolute Error (MAE) and Root Mean Square Error (RMSE) are two common criteria used to evaluate and compare prediction methods (Luo et al., 2022). MAE is used to measure the overall errors of the estimation results and RMSE is used to quantify the stability of the estimation results (W. Zhang et al., 2022). These two criteria are employed as performance metrics for comparison in this study and are defined as below:

$$\text{MAE} = \sqrt{\frac{1}{N}\sum_{k=1}^{N}(\hat{y}(k) - y(k))} \tag{15}$$

$$\text{RMSE} = \sqrt{\frac{1}{N}\sum_{k=1}^{N}(\hat{y}(k) - y(k))^2} \tag{16}$$

where $\hat{y}(k)$ is the observed speed at time $k$ and $y(k)$ is the corresponding predicted traffic speed. $N$ is the size of the testing data set (total number of time intervals).

*4.3. Ridge Regression Results*

Ridge regression was used to interpret the impacts of traffic variables on on-ramp and off-ramp traffic flow estimation in the target domain. Also, the most important variables that being used as inputs for the transfer learning model were selected based on ridge regression results. The coefficients of upstream and downstream traffic variables related to the on-ramp and off-ramp traffic flows were calculated based on the SR51's freeway segments. To identify the most influential variables from the regression model, the coefficient thresholds across a range of values: 0.1, 1, 10, and 100 were examined. Through a comprehensive evaluation of the overall estimation accuracy, it was found that a threshold of 10 provided the optimal criteria. In essence, any variable exhibiting an absolute value surpassing 10 was deemed sufficiently significant to warrant selection as an input variable for the transfer learning model. The outcome of ridge regression analysis is presented in Table 3, which sheds light on the relative importance of each variable. To facilitate easy interpretation, the variables that successfully surpassed the critical threshold of 10 are highlighted in bold-face font. These bold-face variables represent the important few that possess notable influence, signifying their selection as the pivotal components to propel the transfer learning model to the highest traffic flow estimation accuracy.

To reduce randomness, ridge regression was executed 10 times. Hence, the values of the coefficients in Table 3 are the average score of all the 10 regression runs. To better identify the traffic variables, they are named as segment position_traffic variable; the segment positions are named up and down, representing the upstream and downstream.

Table 3 Ridge Regression Results: Selected Variables

| Traffic variables | On-ramp traffic flow | Off-ramp traffic flow |
|---|---|---|
| Up_flow | ***-68.94*** | ***392.27*** |
| Up_flow_variance | ***-66.86*** | ***-66.88*** |
| Up_flow_standard deviation | ***94.42*** | ***94.48*** |
| Up_flow_Kurtosis | -1.20 | -1.20 |
| Up_flow_Skewness | 4.40 | 4.40 |
| Up_speed_mean | 1.60 | 1.61 |
| Up_speed_variance | ***72.12*** | ***72.12*** |
| Up_speed_standard deviation | ***-91.90*** | ***-91.91*** |
| Up_speed_Kurtosis | 0.93 | 0.93 |
| Up_speed_Skewness | 1.98 | 1.98 |
| Up_occupancy_mean | 8.65 | 8.66 |
| Up_occupancy_variance | ***-13.19*** | ***-13.19*** |
| Up_occupancy standard deviation | ***10.82*** | ***10.81*** |
| Up_occupancy_Kurtosis | -0.71 | -0.71 |
| Up_occupancy_Skewness | -1.78 | -1.78 |
| Down_flow | ***106.80*** | ***-371.30*** |
| Down_flow_variance | ***53.07*** | ***53.11*** |
| Down_flow_standard deviation | ***-80.65*** | ***-80.73*** |
| Down_flow_Kurtosis | 0.65 | 0.66 |
| Down_flow_Skewness | 3.66 | 3.66 |
| Down_speed_mean | -1.79 | -1.79 |
| Down_speed_variance | 6.40 | 6.41 |
| Down_speed_standard deviation | -5.15 | -5.17 |
| Down_speed_Kurtosis | 1.15 | 1.15 |
| Down_speed_Skewness | -2.31 | -2.31 |
| Down_occupancy_mean | -2.81 | -2.82 |
| Down_occupancy_variance | 4.34 | 4.34 |
| Down_occupancy standard deviation | -2.74 | -2.73 |
| Down_occupancy_Kurtosis | -0.23 | -0.23 |
| Down_occupancy_Skewness | -5.71 | -5.71 |
| DOW | 0.29 | 0.29 |
| HOD | ***10.28*** | ***10.28*** |
| MOH | -1.14 | -1.13 |

Based on Table 3, it can be observed that MOH and DOW were not selected as important variables for either on or off-ramps models. HOD is selected as an important variable in the on-ramp and off-ramp traffic flow estimation. Intuitively and backed by Table 3, the upstream and downstream mean flows are important variables that influence traffic flow estimation for both on-ramp and off-ramp. Similarly, the variances and standard deviations of upstream traffic variables are also highly related to the on-ramp and off-ramp traffic flow estimation, which signifies that the variations of upstream traffic flow, speed, and occupancy play important roles in the determination of the on-ramp and off-ramp traffic flows. The number of variables representing the upstream traffic state is more than the number of variables representing the downstream traffic state, indicating traffic states in upstream segments affect the estimation of on-ramp and off-ramp flows more than traffic states in downstream segments.

The inclusion of variables from both upstream and downstream traffic ensures that spatial information was considered in the estimation process. Overall, ridge regression ranks the variables that have the highest impact on the on-ramp and off-ramp traffic flow estimation. In addition, it provides an approach to better understanding the interaction between traffic in upstream, downstream, on-ramp, and off-ramp segments.

### 4.4. Traffic Flow Estimation Results

Using the knowledge learned and the model developed for the source domain (SR 51), the missing on-ramp and off-ramp flows for the target domain (L101) were estimated. As it was discussed earlier, to imitate locations where no physical sensors were installed, the on-ramp and off-ramp flows on the target domain (L101) were assumed to be missing. However, the actual data collected from ramps on L101 were utilized as the ground truth data for model performance evaluation. Two measures of effectiveness, MAE and RMSE were used for performance evaluation.

The inputs to the transfer learning model were the variables selected by the ridge regression, and the outputs were on-ramp and off-ramp flows. Table 4 and Table 5 show the estimation performance comparison between four conventional models (KNN, SVR, ANN, and XGB) and the proposed TrA model in terms of MAE. Notably, the results presented in Table 4 and Table 5 are the average values of ten runs of the experiment. The numbers illustrate in Table 4 and Table 5 represent a ramp location with its specific loop number. Based on the results, it can be observed that the proposed TrA model outperforms other models in terms of accuracy in all ramps except 572. Within all the locations, the MAEs for the proposed TrA model range from 23.90 veh/h to 40.85 veh/h for on-ramps, and 31.58 veh/h to 45.31 veh/h for off-ramps.

Table 4 MAE and Training Time Comparison for Estimating the On-ramp Flows

| Model | Loop detector numbering | | | | | | |
|---|---|---|---|---|---|---|---|
| | # 571 | # 572 | # 573 | # 574 | # 575 | # 576 | # 577 |
| KNN | 51.98 | 35.70 | 55.96 | 52.81 | 32.66 | 56.27 | 58.63 |
| SVR | 41.05 | ***30.71*** | 43.44 | 40.60 | 27.03 | 45.85 | 55.84 |
| ANN | 42.54 | 42.04 | 49.56 | 41.58 | 44.96 | 53.10 | 80.00 |
| XGB | 43.30 | 30.93 | 43.98 | 40.58 | ***23.56*** | 43.39 | 45.29 |
| TrA | ***40.85*** | 34.14 | ***29.57*** | ***32.28*** | 28.83 | ***23.90*** | ***36.20*** |

Table 5 MAE and Training Time Comparison for Estimating the Off-ramp Flows

| Model | Loop detector numbering | | | | | | |
|---|---|---|---|---|---|---|---|
| | # 571 | # 572 | # 573 | # 574 | # 575 | # 576 | # 577 |
| KNN | 35.64 | 45.44 | 39.24 | 39.00 | 44.84 | 55.56 | 65.56 |
| SVR | 33.89 | ***30.53*** | 37.52 | 36.62 | 32.47 | 45.87 | 58.81 |
| ANN | 38.73 | 45.91 | 50.08 | 34.27 | 39.78 | 55.84 | 70.39 |
| XGB | 33.44 | 36.81 | 37.49 | 37.98 | 37.07 | 47.46 | 51.23 |
| TrA | ***32.67*** | 34.91 | ***35.00*** | ***35.26*** | ***31.58*** | ***41.91*** | ***45.31*** |

SVR achieves the best estimation performance for both on-ramp and off-ramp of # 572 and XGB outperforms other models when estimating the on-ramp traffic flow for # 575. Figure 4 displays the RMSE values for the flow estimation on the target source. The proposed TrA model acquires the best performance among all the models. Within all the locations, the RMSEs for the proposed TrA model range from 34.55 veh/h to 57.77 veh/h for on-ramps, and 41.75 veh/h to 58.80 veh/h for off-ramps.

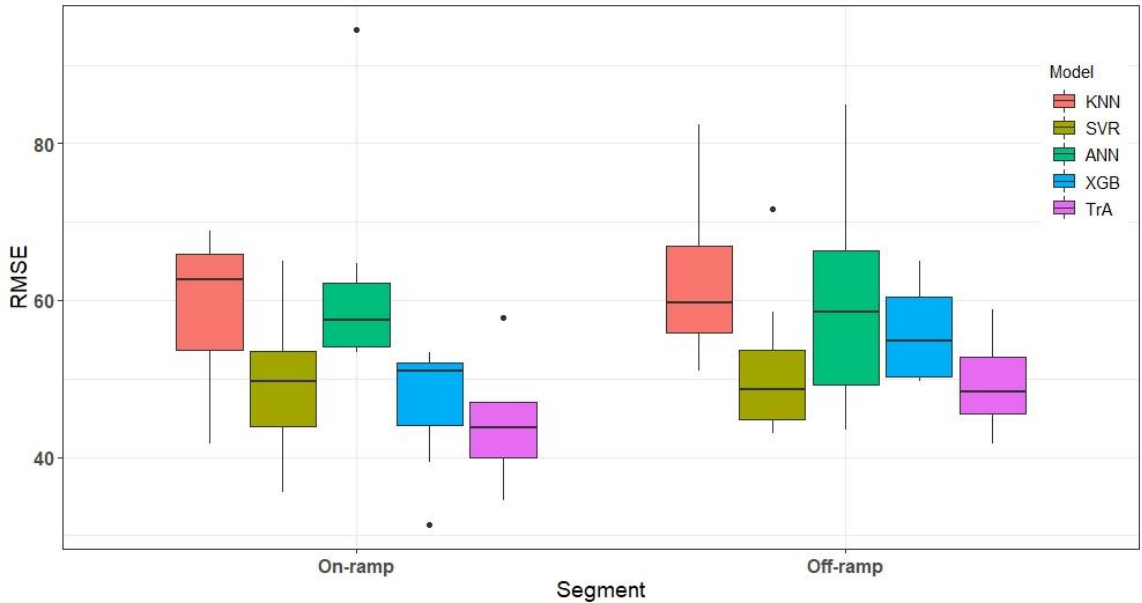

Figure 4 RMSE comparison between different models for different segments

Compared with other conventional machine learning models, TrA has the highest accuracy when estimating both on-ramp and off-ramp traffic flows. This is intuitive, as TrA relaxes the assumption that the data underlying distributions of the source and target domains must be the same. In addition, it handles the estimation of pairs of flows at on-ramp and off-ramp freeway segments with different traffic patterns, distributions, and characteristics. Thus, the proposed TrA model can develop scene-specific models which provide superior estimation accuracy compared to generic models.

When attempting to estimate missing traffic data for on-ramps and off-ramps without physical sensors using ML-based approaches, many proposed estimation models outlined in the literature require extensive data collection. This includes the flow capacity of the weaving area between the on-ramp and off-ramp, on-ramp capacity, and peak on-ramp demand. However, certain data may be unavailable or in need of separate estimation from Highway Capacity Manual guidelines. The need for such extensive data collection may limit the widespread use of proposed estimation models. Fortunately, loop detector data is the only necessary dataset for this research and is commonly used by federal, state, and local agencies for traffic monitoring and operation. This guarantees the proposed approach can be utilized extensively for on-ramp and off-ramp traffic flow estimation.

## 5. Conclusion

For state and local Departments of Transportation to develop the most appropriate control strategy and monitor, maintain, and evaluate the traffic performance of the freeway weaving areas, they need to have access to traffic flows at each pair of on-ramp and off-ramp. However, the ramp flows are not always readily available to transportation agencies and less effort has been done to estimate missing traffic data at on-ramps and off-ramps in locations where no physical sensors have been installed. In this study, an innovative data-driven framework is proposed to estimate the on-ramp and off-ramp traffic flows for locations where no physical sensor is installed.

In the proposed framework, initially, the traffic variables from upstream and downstream of the freeway mainlines were collected from loop detectors. Then, using ridge regression the relationship between the on-ramp and off-ramp traffic flows, and upstream and downstream traffic variables were identified. In addition, the important traffic variables for model development were selected, Next, an inter-corridor similar-location matching algorithm was proposed to match each location of the target corridor to a certain location of the source corridor. Finally, a transfer learning model named TrA was deployed to estimate on-ramp and off-ramp traffic flows at all the locations in the target corridor. The proposed framework was evaluated using freeway traffic data collected from the Southbound stretch of SR51 and L101 located in the Phoenix Metropolitan area, Arizona.

Based on the experimental results, the flow estimation MAEs ranged between 23.90 veh/h to 40.85 veh/h for on-ramps, and 31.58 veh/h to 45.31 veh/h for off-ramps; the flow estimation RMSEs ranged between 34.55 veh/h to 57.77 veh/h for on-ramps, and 41.75 veh/h to 58.80 veh/h for off-ramps. In addition, based on the comparison results, the proposed data-driven model exhibited superior spatial-temporal transferability for new freeway locations compared with traditional machine learning models.

As the framework has high flexibility allowing for the incorporation of multiple variables, future work could consider adding more temporal and spatial variables to the proposed model. Also, the weather conditions, events or incidents, and other factors that might impact the traffic operation can be included in the model to further improve the generalization ability of the model. This paper is the first attempt to employ transfer learning for on-ramp and off-ramp traffic flow estimation. In the future, more advanced machine learning methods and more sophisticated input features could be applied to further improve the estimation performance.